\pdfoutput=1
\documentclass[11pt,twocolumn]{article}

\usepackage[T1]{fontenc}
\usepackage[utf8]{inputenc}
\usepackage{inconsolata}
\usepackage[turkish]{babel}
\usepackage{tabularx}
\usepackage{makecell}
\usepackage{booktabs}
\usepackage{hyperref}
\usepackage{xcolor}
\usepackage{mathptmx}
\usepackage{listings}
\usepackage{microtype}

\makeatletter
\newcommand\notsotiny{\@setfontsize\notsotiny{8.31415}{8.1828}}
\makeatother

\title{Büyük dil modellerinin Türkçe verisetleri ile eğitilmesi ve ince ayarlanması}
\author{A. Taha Arslan}
\date{Mayıs 2023}

\begin{document}

\maketitle

\begin{abstract}
Large language models have advanced enormously, gained vast attraction and are having a phase of intensed research upon them. Some of the developed models and corresponding training datasets have been made public and open-accessible. Hence these may be further fine-tuned with some techniques to obtain specialized models for specific tasks. When it comes to Turkish language, open-access large language models do not provide satisfactory coverage. This can be also observed over published datasets. In this work, we propose some ideas to mitigate this issue. These include creating large Turkish supported datasets, training LLMs with these and fine-tuning already trained models with Turkish inputs. We introduce open-access LLMs and datasets and further report our findings on Turkish-based trainings, problems encountered. We conclude with outcomes of these experiments and propose ideas for further works.
-----
Büyük dil modelleri geçtiğimiz dönemde inanılmaz ölçüde gelişmiş, büyük ilgi toplamış ve üzerlerinde yoğun araştırmaların yapıldığı bir dönem geçirmektedirler. Geliştirilen modeller ve bunları eğitmede kullanılan verisetlerinden bazıları açık erişimli olarak sunulmaktadır. Böylece bunlar üzerinde ince ayarlama teknikleri uygulanarak özelleşmiş görevler için çalışabilir modeller elde etmek mümkündür. Türkçe söz konusu olduğunda sunulan büyük dil modellerinin kapsayıcılığı yeterli düzeyde değildir. Bu durum, yayımlanan verisetlerinde gözlemlenebilir. Bunu gidermenin yolları Türkçe içerikli büyük verisetlerinin oluşturulması, büyük dil modellerinin bunlarla eğitilmesi ve önceden eğitilmiş modellerin Türkçe girdilerle ince ayarlanmaları olarak sayılabilir. Bu çalışmada açık erişimli dil modelleri ve verisetleri üzerinde durulmakta ve Türkçe temelli yapılmış bazı deneyler, karşılaşılan sorunlar ve sonuçlar anlatılmaktadır.
\end{abstract}

\section{Giriş}
Son yıllarda yapay zekâ alanında yaşanan dramatik gelişmelerden birisi de, \emph{büyük dil modellerinin} yani çok sayıda parametreye sahip (bir milyar ve üstünde) yapay sinir ağlarının gene çok sayıda etiklenmemiş metin ve kendi gözetimli ya da yarı gözetimli öğrenim yöntemleri ile eğitilmeleridir. Bu yaklaşım, 2018 yılından itibaren artarak devam etmektedir. Bu gelişmeler ayrıca doğal dil işleme (NLP) araştırmalarının bir önceki yaklaşım olan gözetimli öğrenimden birtakım görevler için farklılaşmarını da beraberinde getirmiştir~\cite{enwiki:largelanguagemodel}.

Büyük dil modelleri genellikle bir cümlede yer alan bir sonraki kelimeyi kestirmek üzere kurgulansalar da, belli görev tanımları ile özelleştirilmeleri ve ince ayarlanmaları sonucunda farklı işlevleri de gerçekleştirebilirler. Bu görevler arasında, \emph{duygu analizi}, \emph{makine tercümesi}, \emph{soru cevaplama}, \emph{eksik kelime tamamlama} bulunmaktadır. İstem teknikleri, modele çözülmek istenen problemi bir metin istemi olarak sunar. Bu yapılırken bu probleme benzeyen bir ya da daha fazlası çözümleriyle birlikte istemin içerisinde yer alabilir. Böylece, dil modeli neyi çözmesi gerektiğini kestirebilir. GPT-3 gibi çok daha güçlü modeller buna gereksinim duymadan bu görevleri yapabilmektedirler.

Son yıllarda geliştirilen önemli büyük dil modellerinin tamamına yakını \emph{transformer} derin öğrenme modelini esas almıştır. Bu modelde en önemli unsur, öz-dikkat denilebilecek (\emph{self-attention}) girdi verinin her bir parçasının önem derecesini ayırt edici biçimde ağırlıklandırma tekniğidir~\cite{enwiki:transformer}. Transformer'ların uygulama alanları arasında makine tercümesi, doküman özetleme, doküman oluşturma, biyolojik dizi analizi ve video anlamlandırma yer almaktadır. Transformer'lar genelde önce gözetimsiz öneğitim ve ardından gözetimli ince ayarlama içeren bir  kendi gözetimli eğitimden geçirilmektedir. GPT-2, GPT-3, GPT-4, BERT, XLNet, RoBERTa ve ChatGPT gibi büyük dil modelleri transformer yapısındadırlar.

Transformer mimarisinin Google tarafından 2017 yılında oluşturulmasının ardından~\cite{vaswani2017attention:transformer}, OpenAI 2018 yılında önceden eğitilmiş üreteç transformer yapısını yayınlamış ve ilk örneği olarak GPT-1 modelini geliştirmiştir~\cite{radford2018improving:gpt1}. Bu yapı 12 öz-dikkat içeren 12 katmandan ve her birinde 64 olmak üzere toplamda 768 adet boyutsal durum içermektedir. Daha sonra geliştirilen GPT-2 modeli, GPT-1'dne göre hem parametre sayısı hem de veriseti bakımından 10 kat büyüktü \cite{radford2019language:gpt2}. GPT-2 kamunun kullanımı için açık erişim olarak yayımlandır. 2020 senesinde sunulan GPT-3 ise 175 milyar parametre ile çok daha büyük bir yapıya sahipti. GPT-3'ün kaynak kodu hiçbir zaman açıklanmadı. OpenAI, GPT-3.5 adı verilen modelin ince ayarlanması ile geliştirilen ve adına ChatGPT dediği ürününü Kasım 2022'de sundu. InstructGPT adı verilen bu ince ayarlamada, eğitilen büyük dil modellerine istemlere yanıt dönme ve istem takip etme özellikleri ekleme amacı güdülmektedir~\cite{openaiblog:instructgpt,ouyang2022training:instructgpt}.

Bu çalışmada, hali hazırda açık erişime sunulmuş bulunan büyük dil modelleri ve bu modellerin eğitilmelerinde ve ince ayarlanmalarında kullanılan gene açık erişimli verisetleri incelenmiştir. Ayrıca, bir büyük dil modelinin Türkçe içerik ile eğitilmesi denenmiş, bir başka önceden eğitilmiş ağın Türkçe istem girdileri ile ince ayarlanması ve bu istemlere yanıt dönmesi incelenmiştir. Bu deneyler ile ilgili yürütülen hazırlıklar ve aşamalar Bölüm~\ref{sec:deneyler}'te, sonuçlar ise Bölüm~\ref{sec:sonuc}'te sunulmuştur.

\begin{table*}[htbp]
{\footnotesize
    \begin{tabularx}{\textwidth}{l l l l}
    \toprule
        Kaynak & Büyüklük & Türkçe içerik & Açıklama\\
        \midrule
        Common Crawl\footnotemark[1] & \makecell[l]{yaklaşık 3.15 milyar web\\sayfası (380 TiB)~\cite{commoncrawl:size}} & \%0.7897~\cite{commoncrawl:languages} & Kâr amacı gütmeyen kuruluş. \\
        \hline
        BookCorpus~\cite{zhu2015aligning:bookcorpus} & 11,000 kitap, 985 milyon kelime. & \emph{bilinmiyor} & \makecell[l]{OpenAI'nın ilk GPT modeli\\için kullanıldı.} \\
        \hline
        C4 ve T5~\cite{2020t5:c4t5} & 745 GB. & Sadece İngilizce veri. & \makecell[l]{Common Crawl verisinin\\temizlenmiş hali. Google tarafından\\yayımlanmıştır.} \\
        \hline
        Openwebtext & 8 milyon doküman, 38GB veri. & Sadece İngilizce veri. & \makecell[l]{GPT-2'in eğitildiği Webtext'e\\alternatif olarak hazırlanmıştır.}\\
        \hline
        RedPajama~\cite{together2023redpajama} & 1.2 Trilyon belirtke. & \emph{bilinmiyor} & \makecell[l]{Diğer büyük kaynaklardan veriseti\\oluşturmayı sağlayan proje.} \\
        \hline
        Vikipedi & 731 MB. & Sadece Türkçe veri. & Bu çalışmada kullanılmıştır. \\
    \bottomrule
    \end{tabularx}
    \caption{Açık erişimli metin kaynakları.}
    \label{tab:opensourcedatasets}
}
\end{table*}

\footnotetext[1]{Common Crawl, belli aralıklarda bu kayıtları almakta ve sunmaktadır. Bu çalışmanın yapıldığı andaki en son arşiv kaydı Mart/Nisan 2023 tarihlidir ve kayıt adı CC-MAIN-2023-14'tür.}

\section{Verisetleri}
\label{sec:verisetleri}
Başarım oranı yüksek bir dil modelinin eğitilebilmesi için gerekli olan en önemli aşamalardan birisi çok büyük ve ön işlemden geçmiş bir metin verisetinin hazırlanmasıdır. Bu aşama, hali hazırda sunulan açık erişimli verisetleri indirilerek yapılabileceği gibi kaynaklar indirilerek sıfırdan da gerçekleştirilebilir. İngilizce ve diğer yaygın diller için bu hazır verisetlerinin kolaylıkla bulunabilmesine karşın malesef Türkçe için hazır verisetleri yeterli ve kolay ulaşılabilir değiller. Bu konudaki bir diğer önemli husus da telif hakları meselesidir. Tablo-\ref{tab:opensourcedatasets}'de açık erişimli veri kaynakları hakkında bilgiler özetlenmiştir.

Hugging Face, Inc. firmasının sağladığı altyapı ile önceden hazırlanmış verisetleri ve derin öğrenme ağ modelleri herkese açık bir şekilde paylaşılmaktadır~\cite{hugginfacedotco}. Bu verisetleri arasında farklı görevler için oluşturulmuş Openwebtext~\cite{Gokaslan2019OpenWebText}, C4 ve PIQA~\cite{Bisk2020PIQA} gösterilebilir. Huggin Face tarafından sunulmakta olan modeller arasında ise farklı görevler için eğitilmiş T5, BERT, BART, GPT-2 ve BLOOM gibi önemli modeller yer almaktadır.

\paragraph{Türkçe metinlerden veriseti oluşturma.} Hali hazırda açık erişimle sunulan verisetlerinde Türkçe içeriğin hiç olmaması ya da çok az yer alması nedeniyle büyük dil ağları araştırmalarında kullanmak üzere sıfırdan bir veriseti oluşturmak elzem olmaktadır. Bu çalışmada yer alan deneyleri yürütülmek için böyle bir veriseti sadece Vikipedi (\emph{Wikipedia} Türkçe sürümü) makaleleri kullanılarak gerçekleştirilmiştir. Bunun için, güncel ve sıkıştırılmış \texttt{trwiki}\footnotemark[2] arşivi indirilmiş ardından bir Python betiği yardımıyla \texttt{json} formatında veriseti oluşturulmuştur. 731 MB büyüklüğündeki bu veride ön işleme ve temizlik yapıldıktan sonra farklı uzunluklarda toplam 818.454 adet metin elde edilmiştir. Bu metinler, \texttt{tiktoken} modülü~\cite{openaigit:tiktoken} ile belirtkeleştirilince 296.1 milyon adet belirtke oluşmuştur. Bu sonuç GPT-2 modelinin de belirtkeleştirildiği 50 bin ögeden oluşan \texttt{r50k\_base} dil kodlaması kullanıldığında elde edilen sayıdır. GPT-3.5 ve GPT-4'te kullanılan 100 bin ögelik \texttt{cl100k\_base} kullanılacak olursa oluşan sayı 242.6 milyon olmaktadır. Bunun Türkçe için daha uygun olduğu düşünülebilir.

\footnotetext[2]{\url{https://dumps.wikimedia.org/trwiki/20230520/}}

\paragraph {Belirtkeleştirme (\emph{Tokenization})} Sözcüksel analizde bir girdi metni oluşturan parçaların sınıflandırılması ve ayırt edilmesi işlemidir. Oluşturulan belirtkeler takip eden bir başka işlemde kullanılırlar. Girdi verisetinden yer alan bütün veri belirtkelere ayrılarak bir sözvarlığı seti oluşturulur. Büyük dil modellerinin eğitilmesinde kullanılan verisetleri üzerinde çoğunlukla \emph{Byte-Pair Encoding} (BPE) belirtkeleştirme algoritması uygulanmaktadır.

\section{Modellerinin eğitilmeleri ve ince ayarlanmaları}
\paragraph {Açık erişimli büyük dil modelleri.} Ticari büyük dil modelleri dışında bazı şahıs ve kurumlar tarafından kaynağı paylaşılan büyük dil modelleri mevcuttur. Bunlar arasında, Meta şirketi tarafından yayımlanan LLaMa~\cite{touvron2023llama} modelinin 7, 13, 33 ve 65 milyar parametre içeren varyantları bulunmaktadır. Bu modeller 1 ve 1.4 trilyon belirtke (\emph{token}) ile eğitilmişlerdir. Malesef eğitim verisetinde yer alan 20 dil içinde Türkçe bulunmamaktadır. BLOOM~\cite{workshop2023bloom} dil modelinde yer alan 46 dil arasında da Türkçe yer almamaktadır.

Kamuya açılmış büyük dil modelleri Tablo-\ref{tab:opensourcemodels}'de özetlenmiştir. Bu modeller eğitilirken çoğunlukla Adam algoritmasının iyileştirilmiş bir versiyonu olan AdamW optimizasyon algoritması kullanılmaktadır~\cite{loshchilov2019decoupled:adamw}.

\paragraph{İnce ayarlama (\emph{fine-tuning}).} Önceden eğitilmiş büyük bir dil modelinin katsayılarının alınarak belli bir görev için, çok daha küçük başka bir verisetiyle tekrar eğitilmesidir. Bu aşama için daha düşük bir öğrenim oranı katsayısı kullanılır. Böylece bu dil modeli bu görev için özelleşmiş olacaktır. Örneğin, istem takip etme bu yeteneklerden biri olabilir. Örneğin, LlaMa modeli açık erişimli ve oldukça büyük olduğu için bu modelin istem takip etme için ince ayarlanması akla yatkın olmaktadır. Bu işlemi Stanford Alpaca gerçekleştirmiştir~\cite{stanfordalpaca}. Çalışmalarında 52,000 adet ve modelin kendiliğinden ürettiği önceden hazırlanmış istem-girdi-cevap ya da istem-cevap şeklinde metin havuzu oluşturmuşlar~\cite{selfinstruct}, LlaMa-7B ve LlaMa-13B modellerini ince ayarlamışlardır. Böylece elde edilen modelin GPT-3.5'a (text-davinci-003) benzer şekilde davrandığını öne sürmektedirler. Buna benzer şekilde alpaca-lora projesi Alpaca sonuçlarını LoRA tekniğini~\cite{hu2021lora} kullanarak çok daha düşük donanım seviyelerinde gerçeklemiştir~\cite{alpacalora}. İki başka proje de, Cabrita ve Zicklein, bu çalışmaları sırasıyla Portekizce'ye ve Almanca'ya taşımışlar yani bu dillerde verilen istemlere ince ayarlanmış LlaMa modeli tarafından yanıt dönülmesini hedeflemişlerdir~\cite{cabrita:por,zicklein:ger}.

\begin{table*}[bthp]
{\footnotesize
    \begin{tabularx}{\textwidth}{l l l l l}
    \toprule
        Model & Sene & Büyüklük & Türkçe içerik & Açıklama \\
        \midrule
        GPT-2 & 2019 & 1.5 milyar parametre. & Yok & OpenAI tarafından. Transformer tabanlı. \\
        \hline
        BLOOM & 2022 & 366 milyar belirtke, 176 milyar parametre. & Yok & Transformer tabanlı. \\
        \hline
        LlaMa & 2023 & 7, 13, 33 ve 65 milyar parametre. & Yok & Meta firması tarafından. Transformer tabanlı.\\
    \bottomrule
    \end{tabularx}
    \caption{Açık erişimli büyük dil modelleri.}
    \label{tab:opensourcemodels}
}
\end{table*}

%\textbf{Dil modellerinin değerlendirilmeleri ve performanslarının karşılaştırılmaları}. Perplexity, MMLU benchmark (Measuring Massive Multitask Language Understanding)

\section{Deneyler}
\label{sec:deneyler}
Bu çalışma kapsamında gerçekleştirilen deneyler iki kısımdan oluşmaktadır. İlkinde, hali hazırda açık erişimli olarak sunulan önceden eğitilmiş büyük dil modelleri ince ayar teknikleri uygulanarak Türkçe istemlere yanıt verir hale getirilmeye çalışılmıştır. İkinci kısımda ise, Türkçe içerikli veriseti oluşturulmuş ve büyük bir dil modeli sıfırdan eğitilmiştir.

\paragraph{Türkçe istemleri takip eden modellerin elde edilmesi.} Bu işlem için, Portekizce ve Almanca için yapılan çalışmalara benzer şekilde, Stanford Alpaca'nın 52.000 adetlik girdisi M2M100 modeli ve belirtkeleştirme aracı~\cite{fan2020englishcentric:m2m100} ile Türkçe'ye çevrilmiştir\cite{hugginface:m2m100}. Bu Türkçeleştirme işlemi oldukça başarılı gözükmektedir. Örneğin ilk iki istem ve yanıtları aşağıdaki şekildedir:

\begin{lstlisting}[breaklines]
{
    "komut":"Sağlıklı kalmak için 3 ipucu verin.",
    "girdi":"",
    "çıktı":"1. Dengeli bir diyet yiyin ve bol miktarda meyve ve sebze içerdiğinizden emin olun. 2. Vücudunuzu aktif ve güçlü tutmak için düzenli olarak egzersiz yapın. 3. Yeterli uyku alın ve tutarlı bir uyku programı tutun."
},
{
    "komut":"Üç ana renk nedir?",
    "girdi":"",
    "çıktı":"Üç ana renk kırmızı, mavi ve sarıdır."
    },
...
\end{lstlisting}

Bir istem, komut ve girdiden oluşabileceği gibi girdi içermeyen bir komuttan da oluşabilir. Dil modeli bu iki türde hazırlanan metinlerden oluşan verisetiyle birkaç tekrarla (\emph{epoch} sayısı 3 alınmıştır) eğitilince buna benzer sorulara yanıt dönebilir hale gelmektedir. LlaMa modellerinden 7 milyar parametre içeren versiyonu bu şekilde eğitilmiştir.

\paragraph{Açık erişimli bir büyük dil modelinin Türkçe verisetiyle eğitilmesi.} Bölüm~\ref{sec:verisetleri}'de anlatıldığı gibi Türkçe içerikli bir veriseti oluşturulmuştur. Açık erişimli olan GPT-2 modeli bu verisetiyle eğitilmiştir. Bu işlem gerçekleştirilirken nanoGPT projesinden~\cite{nanogpt} faydalanılmıştır. Bu modeldeki parametre sayısı 124 milyondur. Veriseti oluşturulurken toplam verinin \%0.1'lik kısmı test veriseti olarak ayrılmış ve bu ayrı tutulan veriseti model eğitimi sırasında aşırı öğrenme ya da eksik öğrenme sorunlarını gözlemek için kullanılmıştır. 8000 iterasyon sonrası eğitim eğitim ve test setleri için sırasıyla \(1.3784\) ve \(1.6127\) kayıp değerlerine ulaşılmıştır. Öğrenim oranı \(0.0006\) ile maksimum olacak şekilde başlatılmış ve iterasyonlar boyunca bir kosinüs fonksiyonu ile azaltılmıştır. Ayrıca belli bir iterasyon sayısı boyunca ısınma eğitimi uygulanmıştır. Bütün çalışmalar bir adet NVIDIA A100-40GB GPU kartı üzerinden yürütülmüştür.

\section{Sonuç ve Tartışma}
\label{sec:sonuc}
LlaMa modelinin Türkçe girdilerle eğitilmemesi neticesinde ilk yapılan çalışmanın sonuçları başarısız olmuştur. Zaten Türkçe kelimeleri görmemiş ve tanımamış bir modelin bu şekilde yanıt dönebilmesini beklemek mantıksız olacaktı. Önümüzdeki çalışmalarda, Türkçe desteği olan ve açık erişimi bulunan büyük bir dil modeli ile bu yöntem tekrarlanabilir.
GPT-2 modelinin eğitiminde elde edilen kayıp oranları nanoGPT tarafından raporlanan sayılardan oldukça düşük çıkmıştır. Bu başta iyi bir şey gibi gözükse de, kullanılan verisetinin küçüklüğü göz önünde bulundurulduğunda bunun altında yatan nedeni ya da nedenleri incelemek faydalı olacaktır. Ayrıca GPT-2 modelinin eğitilmesinde sadece bir metin kaynağından yararlanıldığı için çıktılar tatmin edici ölçüde gerçekleşmemektedir. Bunu geliştirmenin yolu daha büyük ölçüde Türkçe içerikli metinlerden bir veriseti oluşturmak ya da mümkünse böyle bir verisetinin açık erişimli kaynaklardan indirmektir. Elbette büyük dil modellerini büyük verisetleri ile eğitebilmek için daha fazla sayıda donanıma ve GPU kartına erişebilme ihtiyacı da aşikardır.

\par Çalışma boyunca geliştirilen betikler bir Github deposuna yüklenecek ve bu belgenin ilerleyen sürümlerinde paylaşılacaktır.

\bibliographystyle{ieeetr}
\bibliography{references}

\begin{thebibliography}{10}

\bibitem{enwiki:largelanguagemodel}
{Wikipedia contributors}, ``Large language model --- {Wikipedia}{,} the free
  encyclopedia.''
  \url{https://en.wikipedia.org/w/index.php?title=Large_language_model&oldid=1157161819},
  2023.
\newblock [Online; accessed 28-May-2023].

\bibitem{enwiki:transformer}
{Wikipedia contributors}, ``Transformer (machine learning model) ---
  {Wikipedia}{,} the free encyclopedia.''
  \url{https://en.wikipedia.org/w/index.php?title=Transformer_(machine_learning_model)&oldid=1157314320},
  2023.
\newblock [Online; accessed 28-May-2023].

\bibitem{vaswani2017attention:transformer}
A.~Vaswani, N.~Shazeer, N.~Parmar, J.~Uszkoreit, L.~Jones, A.~N. Gomez,
  L.~Kaiser, and I.~Polosukhin, ``Attention is all you need,'' 2017.

\bibitem{radford2018improving:gpt1}
A.~Radford, K.~Narasimhan, T.~Salimans, I.~Sutskever, {\em et~al.}, ``Improving
  language understanding by generative pre-training,'' 2018.

\bibitem{radford2019language:gpt2}
A.~Radford, J.~Wu, R.~Child, D.~Luan, D.~Amodei, I.~Sutskever, {\em et~al.},
  ``Language models are unsupervised multitask learners,'' {\em OpenAI blog},
  vol.~1, no.~8, p.~9, 2019.

\bibitem{openaiblog:instructgpt}
``Openai – aligning language models to follow instructions.''
  \url{https://openai.com/research/instruction-following}.
\newblock Accessed: 2023-05-29.

\bibitem{ouyang2022training:instructgpt}
L.~Ouyang, J.~Wu, X.~Jiang, D.~Almeida, C.~L. Wainwright, P.~Mishkin, C.~Zhang,
  S.~Agarwal, K.~Slama, A.~Ray, J.~Schulman, J.~Hilton, F.~Kelton, L.~Miller,
  M.~Simens, A.~Askell, P.~Welinder, P.~Christiano, J.~Leike, and R.~Lowe,
  ``Training language models to follow instructions with human feedback,''
  2022.

\bibitem{commoncrawl:size}
{Common Crawl}, ``Size of common crawl monthly archives --- statistics of
  common crawl monthly archives.''
  \url{https://commoncrawl.github.io/cc-crawl-statistics/plots/crawlsize},
  2023.
\newblock [Online; accessed 28-May-2023].

\bibitem{commoncrawl:languages}
{Common Crawl}, ``Distribution of languages --- statistics of common crawl
  monthly archives.''
  \url{https://commoncrawl.github.io/cc-crawl-statistics/plots/languages},
  2023.
\newblock [Online; accessed 28-May-2023].

\bibitem{zhu2015aligning:bookcorpus}
Y.~Zhu, R.~Kiros, R.~Zemel, R.~Salakhutdinov, R.~Urtasun, A.~Torralba, and
  S.~Fidler, ``Aligning books and movies: Towards story-like visual
  explanations by watching movies and reading books,'' in {\em Proceedings of
  the IEEE international conference on computer vision}, pp.~19--27, 2015.

\bibitem{2020t5:c4t5}
C.~Raffel, N.~Shazeer, A.~Roberts, K.~Lee, S.~Narang, M.~Matena, Y.~Zhou,
  W.~Li, and P.~J. Liu, ``Exploring the limits of transfer learning with a
  unified text-to-text transformer,'' {\em Journal of Machine Learning
  Research}, vol.~21, no.~140, pp.~1--67, 2020.

\bibitem{together2023redpajama}
T.~Computer, ``Redpajama: An open source recipe to reproduce llama training
  dataset.'' \url{https://github.com/togethercomputer/RedPajama-Data}, April
  2023.

\bibitem{hugginfacedotco}
``Hugging face – the ai community building the future..''
  \url{https://huggingface.co/}.
\newblock Accessed: 2023-05-28.

\bibitem{Gokaslan2019OpenWebText}
E.~P. S.~T. Aaron~Gokaslan, Vanya~Cohen, ``Openwebtext corpus.''
  \url{http://Skylion007.github.io/OpenWebTextCorpus}, 2019.

\bibitem{Bisk2020PIQA}
Y.~Bisk, R.~Zellers, R.~L. Bras, J.~Gao, and Y.~Choi, ``Piqa: Reasoning about
  physical commonsense in natural language,'' in {\em Thirty-Fourth AAAI
  Conference on Artificial Intelligence}, 2020.

\bibitem{openaigit:tiktoken}
``Openai – how to count tokens with tiktoken.''
  \url{https://github.com/openai/openai-cookbook/blob/main/examples/How_to_count_tokens_with_tiktoken.ipynb}.
\newblock Accessed: 2023-05-29.

\bibitem{touvron2023llama}
H.~Touvron, T.~Lavril, G.~Izacard, X.~Martinet, M.-A. Lachaux, T.~Lacroix,
  B.~Rozière, N.~Goyal, E.~Hambro, F.~Azhar, A.~Rodriguez, A.~Joulin,
  E.~Grave, and G.~Lample, ``Llama: Open and efficient foundation language
  models,'' 2023.

\bibitem{workshop2023bloom}
T.~L. Scao and et~al., ``Bloom: A 176b-parameter open-access multilingual
  language model,'' 2023.

\bibitem{loshchilov2019decoupled:adamw}
I.~Loshchilov and F.~Hutter, ``Decoupled weight decay regularization,'' 2019.

\bibitem{stanfordalpaca}
R.~Taori, I.~Gulrajani, T.~Zhang, Y.~Dubois, X.~Li, C.~Guestrin, P.~Liang, and
  T.~B. Hashimoto, ``Stanford alpaca: An instruction-following llama model.''
  \url{https://github.com/tatsu-lab/stanford_alpaca}, 2023.

\bibitem{selfinstruct}
Y.~Wang, Y.~Kordi, S.~Mishra, A.~Liu, N.~A. Smith, D.~Khashabi, and
  H.~Hajishirzi, ``Self-instruct: Aligning language model with self generated
  instructions,'' 2022.

\bibitem{hu2021lora}
E.~J. Hu, Y.~Shen, P.~Wallis, Z.~Allen-Zhu, Y.~Li, S.~Wang, L.~Wang, and
  W.~Chen, ``Lora: Low-rank adaptation of large language models,'' 2021.

\bibitem{alpacalora}
``alpaca-lora: Instruct-tune llama on consumer hardware.''
  \url{https://github.com/tloen/alpaca-lora}, 2023.

\bibitem{cabrita:por}
``Cabrita: A portuguese finetuned instruction llama.''
  \url{https://github.com/22-hours/cabrita}, 2023.

\bibitem{zicklein:ger}
``Zicklein: A german finetuned instruction llama.''
  \url{https://github.com/avocardio/Zicklein}, 2023.

\bibitem{fan2020englishcentric:m2m100}
A.~Fan, S.~Bhosale, H.~Schwenk, Z.~Ma, A.~El-Kishky, S.~Goyal, M.~Baines,
  O.~Celebi, G.~Wenzek, V.~Chaudhary, N.~Goyal, T.~Birch, V.~Liptchinsky,
  S.~Edunov, E.~Grave, M.~Auli, and A.~Joulin, ``Beyond english-centric
  multilingual machine translation,'' 2020.

\bibitem{hugginface:m2m100}
``M2m100.''
  \url{https://huggingface.co/transformers/v4.4.2/model_doc/m2m_100.html}.
\newblock Accessed: 2023-05-29.

\bibitem{nanogpt}
``nanogpt.'' \url{https://github.com/karpathy/nanoGPT}, 2023.

\end{thebibliography}

\end{document}